%% file: main.tex
\documentclass[letterpaper, 10 pt, journal, twoside]{IEEEtran}

\DeclareUnicodeCharacter{2212}{\ensuremath{-}}
\usepackage{graphicx}
\usepackage{booktabs}
\usepackage{amsmath}
\usepackage{xcolor} 
\usepackage{bm}
\usepackage{wrapfig, enumitem}
\usepackage{multirow}
\usepackage{diagbox}
\usepackage{url}

\newcommand{\rebuttal}[1]{\textcolor{black}{#1}}

\newcommand{\algo}{Force-Constrained Visual Policy}

\begin{document}

\title{
Force-Constrained Visual Policy: Safe Robot-Assisted Dressing via Multi-Modal Sensing
}

\author{Zhanyi Sun$^*$, Yufei Wang$^{*}$, David Held$^\dag$, Zackory Erickson$^\dag$
\thanks{Manuscript received: Oct, 11th, 2023; Revised January, 18th, 2024; Accepted March, 7th, 2024.}
\thanks{This paper was recommended for publication by Editor Angelika Peer upon evaluation of the Associate Editor and Reviewers’ comments. This work was supported by the National Science Foundation under Grant No. IIS-2046491.}
\thanks{*Equal Contribution. $^\dag$Equal Advising. }
\thanks{All authors are with the Robotics Institute, Carnegie Mellon University. Emails: \{zhanyis, yufeiw2, dheld, zerickso\}@andrew.cmu.edu }
\thanks{Digital Object Identifier (DOI): see top of this page.}
}

\markboth{IEEE Robotics and Automation Letters. Preprint Version. Accepted March, 2024}
{Sun \MakeLowercase{\textit{et al.}}: FCVP: Safe Robot-Assisted Dressing via Multi-Modal Sensing}

\maketitle

\begin{abstract}

Robot-assisted dressing could profoundly enhance the quality of life of adults with physical disabilities. To achieve this, a robot can benefit from both visual and force sensing. The former enables the robot to ascertain human body pose and garment deformations, while the latter helps maintain safety and comfort during the dressing process. In this paper, we introduce a new technique that leverages both vision and force modalities for this assistive task. Our approach first trains a vision-based dressing policy using reinforcement learning in simulation with varying body sizes, poses, and types of garments.
We then learn a force dynamics model for action planning to ensure safety. Due to limitations of simulating accurate force data when deformable garments interact with the human body, we learn a force dynamics model directly from real-world data. Our proposed method combines the vision-based policy,  trained in simulation, with the force dynamics model, learned in the real world, by solving a constrained optimization problem to infer actions that facilitate the dressing process without applying excessive force on the person. We evaluate our system in simulation and in a real-world human study with 10 participants across 240 dressing trials, showing it greatly outperforms prior baselines. Video demonstrations are available on our project website\footnote{\url{https://sites.google.com/view/dressing-fcvp}}.

\end{abstract}



\input{sections/intro}

\input{sections/related_work}
\input{sections/problem_def}
\input{sections/method}
\input{sections/experiments}

\input{sections/conclusion}


\vspace{-0.08in}
\bibliographystyle{IEEEtran}
\bibliography{IEEEabrv}

\end{document}

%% file: sections/intro.tex
\section{Introduction}

\IEEEPARstart{D}{ressing} is a crucial activity for individuals with disabilities or limited mobility to receive assistance with. Recent studies~\cite{harris2019long} estimate that 92\% of all residents in nursing facilities and at-home care patients require assistance with dressing. Robot-assisted dressing has emerged as a potential solution to these challenges~\cite{Wang2023One, zhang2022learning, kapusta2019personalized, gao2016iterative}, which could be used to enhance the  quality of life of people with physical disabilities. In this work, we demonstrate a new learning-based method for combining vision and force sensing modalities towards a safe and comfortable assistive dressing system. 

Robot-assisted dressing comes with several challenges. 
First, robotic manipulation of deformable garments is challenging due to the lack of a compact  state space representation, complex cloth dynamics, and self-occlusions of clothing. 
Moreover, during robot-assisted dressing, the robot works in proximity to the human and has direct physical contact with the human body. Undesired motions performed by the robot that strain the garment or cause accidental collisions with a person could apply large forces to the human body and pose discomfort and potential safety risks.

Prior work in cloth manipulation and robot-assisted dressing has demonstrated the use of vision~\cite{Wang2023One, gao2015user} and force~\cite{erickson2018deep} modalities separately to make control decisions. Yet, there is a clear advantage to leveraging both modalities 
simultaneously~\cite{zhang2019probabilistic}. 
Visual sensing is useful to observe the garment and human arm to infer a reasonable dressing path, and force sensing is needed to ensure safety and comfort during the dressing process. In this context, simulation can be used to collect large amounts of data to train a control policy that can generalize across diverse people, body poses, and garments. Prior work~\cite{Wang2023One} has demonstrated the ability to transfer point cloud-based assistive dressing policies from simulation to the real world; however, most simulators do not provide sufficiently accurate robot force sensing when manipulating deformable garments around human bodies, which limits the transfer of force-based models from simulation to the real world.
\begin{figure}[t]
\centering
\includegraphics[width=0.5\textwidth]{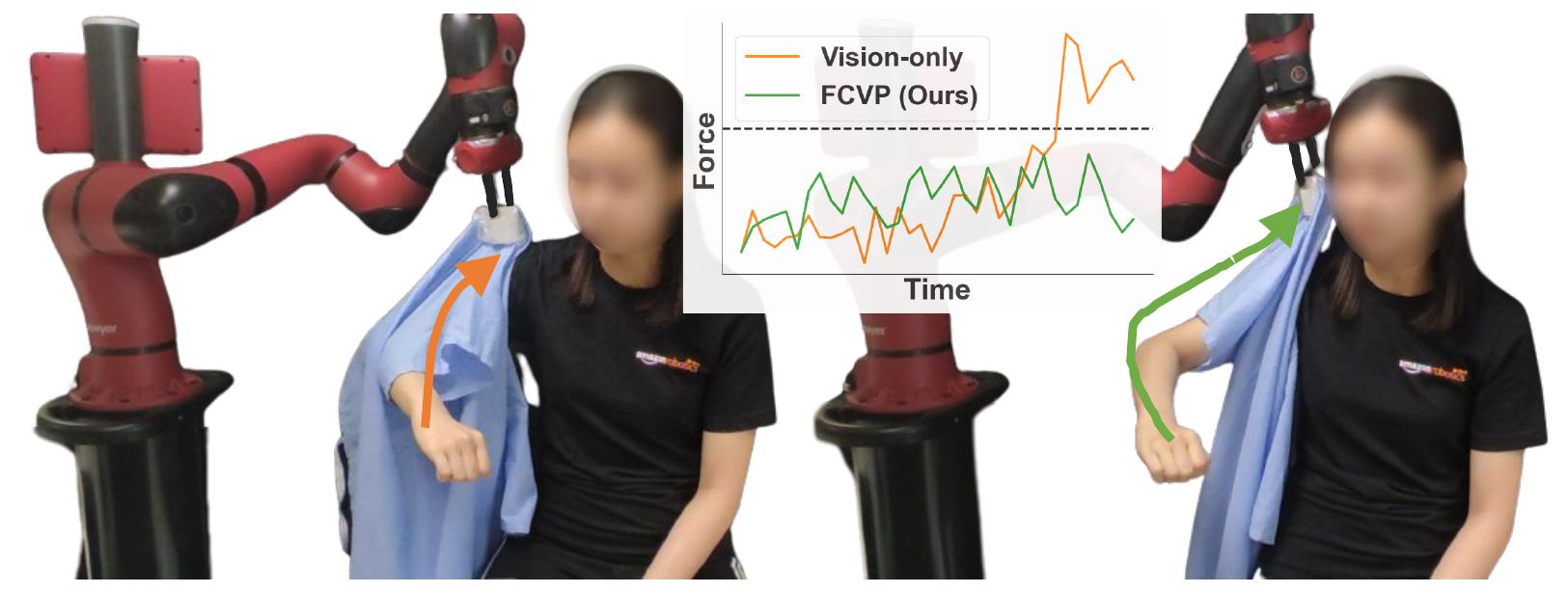}
  \caption{Our method learns a force dynamics model in the real world to constrain the vision-based policy trained in simulation (right),  preventing high force from being applied to the person (left). 
  }
  \label{fig:intro_example}
  \vspace{-0.28in}
\end{figure}
This reality gap necessitates learning from force measurements directly in the real world. The question that we explore in this paper is how to combine a visual policy for assistive dressing trained in simulation with force sensing data that is only available in the real world.

In this paper, we propose a new method for the task of assistive dressing, named Force-Constrained Visual Policy (FCVP), shown in Figure~\ref{fig:intro_example}. Our method elegantly handles the case in which only the visual modality (using point clouds) can be simulated sufficiently accurate to be transferred to the real world, but the force modality cannot.
Our key idea is to use a vision-based policy trained in simulation to propose actions, and then to use a force-based dynamics model trained in the real world to filter out unsafe actions.
We comprehensively evaluate our method with a real-world human study with 10 participants and 240 trials, demonstrating its strong effectiveness. 

In summary, we make the following contributions:
\begin{itemize}[itemsep=0pt,topsep=0pt]
\item We propose a new method for multi-modal learning when one sensor cannot be well-modeled in simulation.  Our method, \algo~(FCVP),  combines a vision-based policy trained in simulation with a force-based dynamics model trained in the real world.
\item We evaluate our method rigorously in simulation and also perform a real-world human study with 10 participants and 240 dressing trials to quantify the real-world practicality and efficacy of the proposed method and system. These experiments demonstrate that our method leads to a safe and comfortable assistive dressing system with higher dressing performance by ensuring low forces are exerted to the human.
\end{itemize}

%% file: sections/related_work.tex
\section{Related Work}
\textbf{Robot Assisted Dressing:} 
A large body of works have studied the problem of robot-assisted dressing~\cite{Wang2023One, zhang2022learning}, and some of those have investigated how to use the force modality to minimize force during the dressing task~\cite{gao2016iterative, erickson2018deep,zhang2019probabilistic, zhang2017personalized}. 
Visual inputs are either not used in these works~\cite{erickson2018deep}, or simply used for detecting initial waypoints on the human arm for interpolating a dressing path~\cite{zhang2019probabilistic,zhang2017personalized}. Our method differs from these approaches in that we 
learn a force dynamics model to filter actions proposed from a vision policy. Some other works~\cite{Wang2023One,zhang2020learning,tamei2011reinforcement,koganti2014real,pignat2017learning} leverage only vision to perform a dressing task with no force sensing. 
Ours differ from those as we leverage both vision and force to ensure safety and comfort during dressing. 


Another line of work related to ours studies force perception and simulation during robot-assisted dressing~\cite{erickson2017does, yu2017haptic, kapusta2016data, wang2022visual}. Some of these papers study force sensing during robot-assisted dressing in simulation~\cite{erickson2017does, wang2022visual} without a quantitative real-world verification. Others~\cite{yu2017haptic} show that with system identification, simulation parameters can be tuned to approximate robot force sensing when dressing a known garment on a fixed human body pose, yet large error still remains when the garment undergoes deformations not presented in the system identification data. Our work differs from these, as instead of tuning the simulation parameters for force sensing and performing sim2real transfer, we directly learn a force dynamics model in the real world. 



\textbf{Multi-Modal Learning for Robotic Manipulation:} 
Recently, there has been an increasing number of works studying multi-modal learning, which combines modalities such as vision and force~\cite{zhang2019probabilistic}, vision and touch~\cite{sunil2023visuotactile,lee2019making, watkins2019multi, calandra2018more, wi2022virdo, sundaresan2022learning}, and vision and audio~\cite{du2022play, yang2017deep, li2022see}, with applications in grasping~\cite{watkins2019multi, calandra2018more}, object manipulation~\cite{sunil2023visuotactile, wi2022virdo,li2022see, beltran2020variable}, assistive tasks~\cite{zhang2019probabilistic, zhang2017personalized, sundaresan2022learning}, and many more. 
Most prior works on multimodal learning focus on how to design a more effective policy network architecture that takes all modalities as input. Most of them directly train in the real world~\cite{lee2019making, calandra2018more, sundaresan2022learning, du2022play, li2022see}, via imitation learning~\cite{sundaresan2022learning, du2022play, li2022see}, supervised learning~\cite{calandra2018more}, or self-supervised learning followed by reinforcement learning~\cite{lee2019making}, which all require a large amount of human-collected datasets or robot trials. In contrast, our approach trains the vision-based policy in simulation.  Some approaches train a multimodal policy in simulation and perform sim2real transfer, such as vision with contact points~\cite{watkins2019multi}, and vision with rigid-body force sensing~\cite{beltran2020variable}. These approaches assume that all of the modalities can be accurately simulated; in contrast, our method trains a force-based dynamics model directly in the real world, without assuming that the force modality can be accurately simulated for modeling cloth-human force interactions.
In contrast to these prior works which employ a single policy network that handles both modalities, our proposed approach combines a vision-based policy trained in simulation and a force dynamics model trained in the real-world via solving a constrained optimization problem.


\textbf{Safe Reinforcement Learning}:  The objective of developing a safe robot-assisted dressing system can be formulated as a safe reinforcement learning problem~\cite{garcia2015comprehensive, liu2022constrained}, where the reward is to dress the person, and the safety constraint is that the amount of  force applied to the person should be below a threshold. However, there could be some issues of directly applying safe RL algorithms to our problem setting: if we train the safe RL policy in simulation using both vision and force, then it will be difficult to transfer to the real world since the simulator does not provide accurate enough force simulation with deformable cloth. Our method avoids this issue by training a vision-based policy in simulation and a force-based dynamics model in the real world.  

%% file: sections/problem_def.tex
\section{Problem Statement and Assumptions }
As shown in Figure~\ref{fig:intro_example}, we study the task of single-arm dressing assistance, where the goal is to dress the sleeve of a garment onto a person's arm.  Single-arm assistive dressing is a fundamental skill of full upper-body dressing assistance for individuals with motor impairments. We want to achieve safe dressing assistance by ensuring that low forces are exerted to the human during the dressing process.
Formally, let $f_t \in \mathcal{R}$ be the amount of force the garment exerts onto the human body at time step $t$ (which we approximate through force sensing at the robot's end-effector). 
We want to develop a safe robot-assisted dressing system that can pull the garment to cover the human arm and shoulder, while maintaining the  force applied to the human to be below a threshold  $\tau$, i.e., $f_t < \tau$, $\forall t\in [1,T]$, where $T$ is the horizon of the task.
We assume the person holds their arm static during the dressing process, and that the robot has already grasped the opening of the garment shoulder in preparation for dressing. While not the focus of this paper, prior works have proposed methods for grasping garments~\cite{zhang2022learning, zhang2020learning} and adapting to human motion during dressing assistance~\cite{erickson2022characterizing}, which could be integrated into our work.


\section{Background - Vision-based Policy Learning in simulation}
\label{sec:vision-learning}
Our method leverages a vision-based policy $\pi^v$ trained in simulation from prior work~\cite{Wang2023One}. 
We describe the core training procedure here, and refer to~\cite{Wang2023One} for more details.
The dressing task is formulated as a Partially Observable Markovian Decision Process (POMDP) 
and is solved via reinforcement learning. The core components of the POMDP are defined as follows:

\noindent\textbf{Observation Space $O$}: The policy observation is the segmented point cloud of the scene, which consists of the garment point cloud $P^g$ and the human arm point cloud $P^h$. As we assume the human to be static, we can obtain the static arm point cloud $P^h$ before the garment occludes the arm and use it during the whole dressing process; thus our input includes the full arm even when the garment occludes it. 
A single point $P^r$ at the location of the robot's end-effector is added to the observation. The full observation $o$ is the concatenation of these three point clouds: $o = [P^g; P^h; P^r]$ (see Figure~\ref{fig:system} for a visualization). The feature for each point in $o$ is a one-hot encoding indicating which object the point belongs to, i.e., the garment, the human arm, or the end-effector.  

\noindent\textbf{Action Space $A$}: the action $a \in A$ is defined as the delta transformation for the robot end-effector. It is a 6D vector, where the first 3 elements denote the delta translation, and the second 3 denote the delta rotation represented as axis angle. 

\noindent\textbf{Reward $r$}: 
The reward $r$ consists of a term that measures the task progress, which is the dressed distance of the garment along the human arm, with additional auxiliary reward terms to prevent the gripper from moving too close to the person. The full detailed reward function is the same as in Wang et al.~\cite{Wang2023One}, and can be found on our project website. 

SAC~\cite{haarnoja2018soft} is used as the underlying RL algorithm for training the vision-based policy,
and a segmentation-type PointNet++~\cite{qi2017pointnet++} as the policy architecture (see \cite{Wang2023One} for details). 
As in prior work~\cite{Wang2023One}, the vision-based policy is trained in simulation on many variations of body shapes, arm poses, and garments, and can be transferred to a real world manipulator. However, the actions inferred by the vision-based policy may exert high forces onto people when deployed in the real world.
Our method handles this by further learning a force dynamics model in the real world.

%% file: sections/method.tex
\begin{figure*}
    \centering
    \includegraphics[width=.75\textwidth]{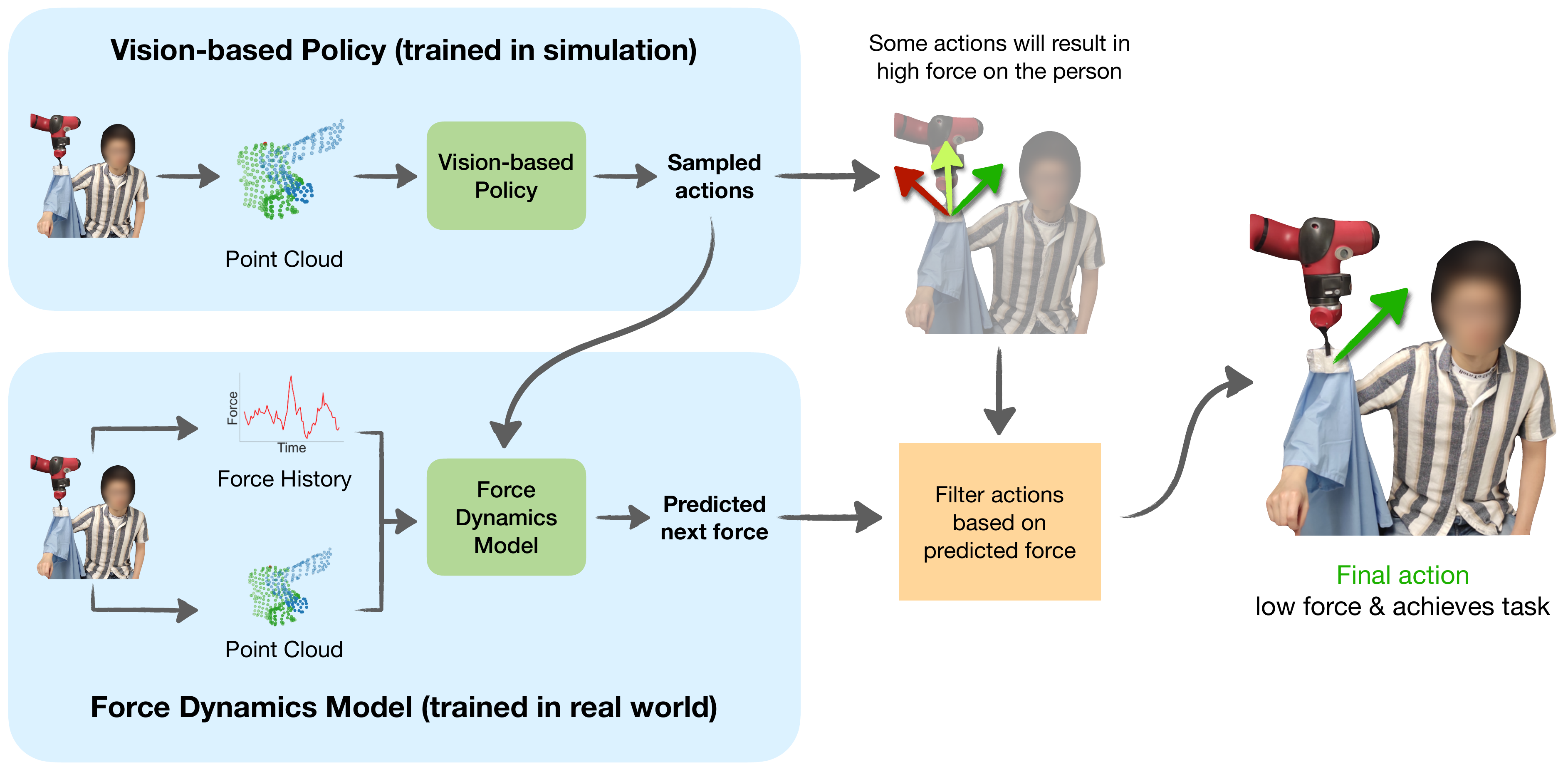}
    \vspace{-0.08in}
    \caption{Our system combines a vision-based policy and a force dynamics model to achieve safe robot-assisted dressing.
    As most simulators provide sufficiently accurate simulation of point clouds yet not the force modality for sim2real transfer, the vision-based policy is trained with a large amount of data in simulation, and the force dynamics model is trained with a small amount of data in the real world. 
    At test time, the vision-based policy proposes action samples that progress the dressing task. The force dynamics model predicts the future forces of these sampled actions, and the predictions are used to filter actions that are unsafe, i.e., those applying too much force to the human. The final chosen action is safe with low force and achieves the task.
    }
    \vspace{-0.15in}
    \label{fig:system}
\end{figure*}

\section{Method}

\textbf{Method Overview:}  
As shown in Figure~\ref{fig:system}, our system is comprised of two parts.  First, we leverage a vision-based policy from prior work which is trained in simulation using reinforcement learning (as described in section~\ref{sec:vision-learning}). By using simulation, we are able to easily collect a large amount of data and train a single policy that can generalize to many variations of human arm poses, body shapes, and garments. The vast amount of data needed makes it prohibitive to train the vision-based policy directly in the real world. To ensure safe assistive dressing, we learn a force dynamics model which predicts the future forces applied to the human. The force dynamics model is trained in the real world, due to the fact that many simulators do not provide sufficiently accurate force simulation for deformables manipulated around the human body. 
At test time, the final robot action is inferred by solving a constrained optimization problem that combines the vision-based policy and the force dynamics model.

\subsection{Force dynamics model learning in the real world}
\label{sec:learning_force}
As previously mentioned, force simulation of deformable garments is not sufficiently accurate to transfer from simulation to the real world. 
Even after system identification, it is challenging to accurately estimate all the local forces caused by cloth deformation and stretch when interacting with a human or other objects in the environment. 
Therefore, we aim to directly learn the force-based model in the real world. 

Specifically, we learn a force dynamics model $d_\psi(o_t, F_t, a_t)$, which takes as input the current point cloud observation $o_t$ as described in section~\ref{sec:vision-learning}, the past $N$ steps of forces $F_t = [\tilde{f}_t, ..., \tilde{f}_{t-N+1}]$ (where $\tilde{f} \in \mathcal{R}^3$ is a three-dimensional force vector), and the robot action $a_t$.  The force dynamics model predicts the future amount of force $\hat{f}_{t+1} \in \mathcal{R}$ the human will experience due to robot action $a_t$.
To predict the future force, the force dynamics model uses a PointNet++~\cite{qi2017pointnet} encoder to encode the point cloud observation $o_t$ into a latent vector. This latent vector is then concatenated with the force history $F_t$ and the action $a_t$. Another MLP then receives as input the concatenated vector and outputs the predicted force $\hat{f}_{t+1}$ in the next timestep.

We collect training data for the force dynamics model in the real world using the vision-based policy $\pi^v$. 
To train the force dynamics model on a wider range of action distributions, 
at each time step, the action $a_t$ is sampled as following: with probability $p$, $a_t$ is uniformly sampled from $[-1, 1]^{|A|}$ (where $|A|$ is the dimension of the action space), and with probability $1-p$, it is sampled from $a_t \sim \pi^v(o_t)$. 
The force dynamics model (including the PointNet++ encoder and the MLP) is trained using the MSE loss to predict the future force: $L(\psi) = ||d_\psi(o_t, F_t, a_t) - f_{t+1}||_2^2$, where $\psi$ denotes the parameters for the force dynamics model. 


\subsection{Force-Constrained Vision Policy } 
At test time, we combine the trained vision policy $\pi^v$ and the force dynamics model $d_\psi$
to infer the action by solving the following constrained optimization problem:
{\small
\begin{equation}
    \arg\max_a  ~~\pi^v(a | o) \quad 
    \text{subject to}  ~~d_\psi(o, F, a) \leq \tau
\end{equation}
}
where $\tau$ is a user-chosen force threshold. 
There are several optimization algorithms for solving such a constrained optimization problem, such as Lagrangian method, active-set method, interior-point method, random shooting method, and more. As the functions involved in our constrained optimization problem are represented using neural networks, we use the random shooting method to solve the optimization problem due to its simplicity and low computational cost. The random shooting method works as follows: we sample a set of actions, filter out any actions whose predicted forces are above the threshold $\tau$, and among the actions whose predictions are below the threshold, we execute the action with the highest probability under the vision-based policy $\pi^v$. We use the same action sampling distribution as during training (i.e. a mix of actions from the vision policy $\pi^v$ and actions from a uniform distribution). 
If there is no action whose predicted force is below the threshold, we execute the action with the lowest predicted force. 



%% file: sections/experiments.tex
\begin{figure*}[t]
    \centering
    \begin{minipage}{0.35\textwidth} 
        \includegraphics[width=\linewidth]{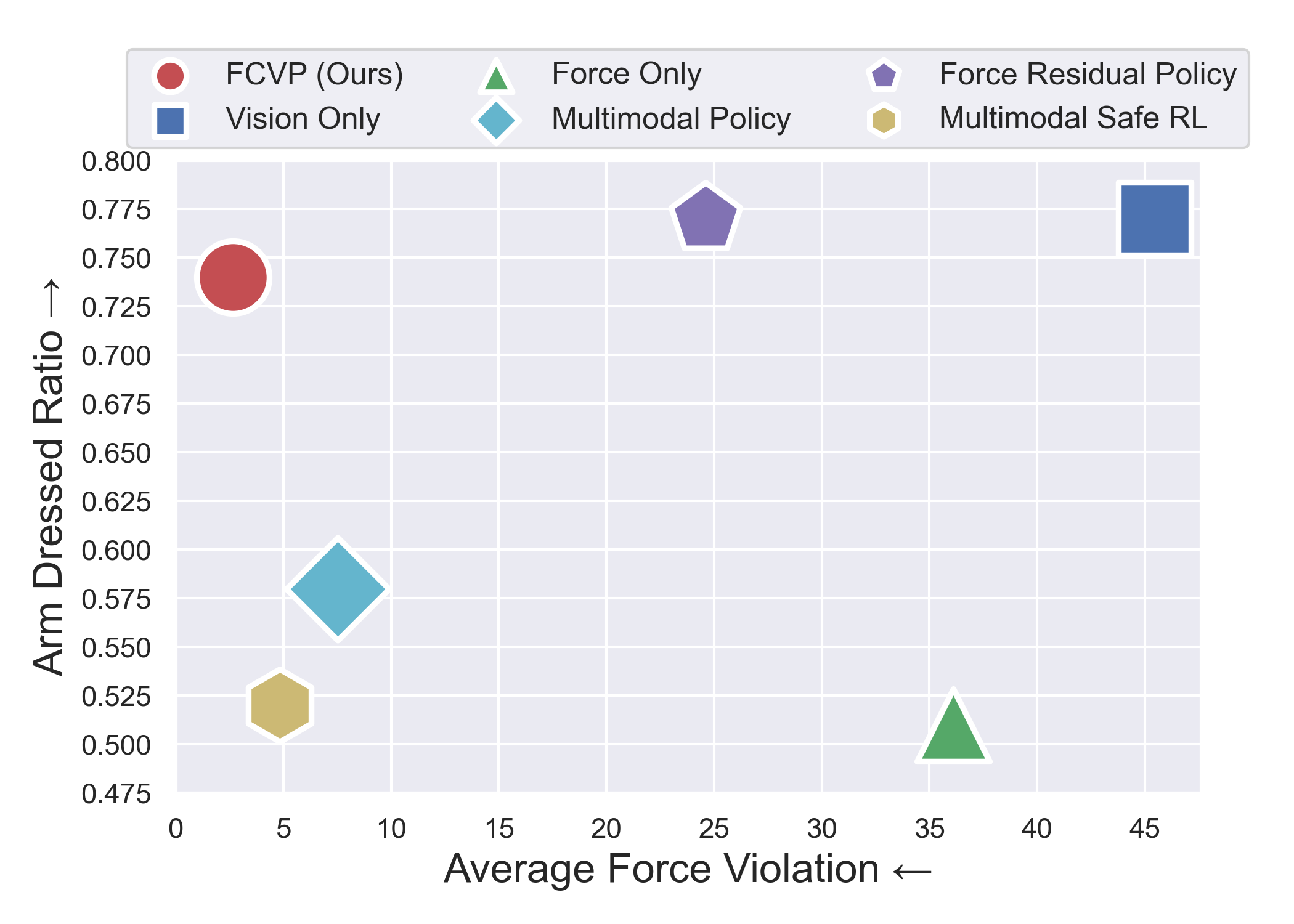}
    \end{minipage}
    \hfill 
    \begin{minipage}{0.6\textwidth} 
        \scriptsize
        \begin{tabular}{cccc}
        \toprule
        & \multicolumn{1}{c}{\begin{tabular}[c]{@{}c@{}} Arm \\ Dressed Ratio $\uparrow$  \end{tabular}}  & \multicolumn{1}{c}{\begin{tabular}[c]{@{}c@{}}Average Force \\Violation $\downarrow$   \end{tabular}}  & \multicolumn{1}{c}{\begin{tabular}[c]{@{}c@{}} \# Training trajectories \\ in sim B $\downarrow$ \end{tabular}} \\ \midrule
        FCVP (Ours)  &   $0.74 \pm 0.29 $ & $\mathbf{2.65}$ &  225\\ 
        Vision Only  &  $\mathbf{0.77 \pm 0.33}$ & $45.45$ & \textbf{0} \\ 
        Force Only  &   $0.51 \pm 0.28$ & $36.10$  & 225 \\ 
        Multimodal Policy  &  $0.58 \pm 0.42$ & $7.51$ & 7987 \\
        Force Residual Policy & $\mathbf{0.77 \pm 0.25}$ & $24.60$  & 8190\\  
        Multimodal Safe RL  & $0.52 \pm 0.40$ & $4.83$  & 8278\\ 
        \bottomrule
        \end{tabular}
    \end{minipage}
    \vspace{-0.05in}
    \caption{
    (Left) Among all compared methods, FCVP achieves the best trade-off between the arm dressed ratio and the force violation amount. (Right) The detailed quantitative results for each method, as well as the number of training trajectories required for convergence in sim B.
    }
    \label{fig:sim2sim_results}
    \vspace{-0.2in}
\end{figure*}









\section{Experiments}
We conduct both simulation and real-world experiments to evaluate our method.
We first perform sim2sim transfer experiments to compare FCVP with other multimodal learning methods (Section~\ref{sec:sim2sim_experiments}). We also perform real-world human studies (Carnegie Mellon University IRB Approval under 2022.00000156) to evaluate the effectiveness of the robot-assisted dressing system (Section~\ref{sec:human_study}).  

\subsection{Sim2sim Transfer Experiments}
\label{sec:sim2sim_experiments}

\noindent\textbf{Setup: }
In order to test our method in a controlled setting, we create two simulation environments with different simulation parameters (the detailed parameters can be found on our project website);  we treat one of them as simulation (sim A), and the other as an approximation to the real world (sim B). The force readings between these two environments are different due to the differences in the simulation parameters, approximating the sim2real gap. We use NVIDIA FleX~\cite{macklin2019non} wrapped in SoftGym~\cite{lin2021softgym} as the simulator. 

We use SMPL-X~\cite{SMPL-X:2019} to generate human meshes with distinct body shapes, sizes, and arm poses.
Specifically, we generate 4 different arm pose regions;  the arm poses within each region are similar to each other with small variations, and the arm poses are very different across regions (see figures of the arm pose regions on our project website). For each region, we generate 45 human meshes of distinct body shapes and sizes. All methods are evaluated on each arm pose region, and then the results are averaged across the arm pose regions. 
We use the same garments as in Wang et al.~\cite{Wang2023One}: a common hospital gown and 4 cardigans from the Cloth3D dataset~\cite{bertiche2020cloth3d}. 
In simulation, we set the force threshold $\tau$ to be 40 units (note due to simulation inaccuracies, this unit does not correspond to Newtons in the real world.)
We compare the following methods: 

\textbf{FCVP (Ours):} For our method FCVP, we learn a  vision-based policy in sim A. The force dynamics model is learned in sim B. We collect one trajectory per human mesh and garment, resulting in a total of 225 trajectories in each arm pose region for training the force dynamics model in sim B.  
We use the past 5 steps of force measurements as input to the force dynamics model. 

\textbf{Vision Only} directly transfers the vision-based policy trained in sim A to sim B, without any fine-tuning nor using the force information. 

\textbf{Force Only} uses only the force dynamics model trained in sim B, without using the vision-based policy pretrained in sim A. The actions are planned by minimizing the predicted force, and are heuristically sampled within a forward task progression cone, similar to the method used in~\cite{erickson2018deep}. 

\textbf{Multimodal Policy:} We first pretrain a policy that takes as input the vision and the force modality in sim A using RL, and then we fine-tune it with vision and force in sim B. This is the most standard approach for multimodal learning. 

\textbf{Force Residual Policy:} We first pretrain a policy that only uses the vision modality in sim A, and then we train a residual policy on top of this pretrained vision-based policy using both vision and force in sim B. The residual policy is trained to output a delta action, which is added to the action from the pretrained vision-based policy. 

\textbf{Multimodal Safe RL:} This is similar to Multimodal Policy, but instead of using RL to pretrain or fine-tune the policy, we use safe-RL. We use SAC-Lagrangian~\cite{liu2022constrained} as the training algorithm. Specifically, we first pretrain a policy that takes as input the vision and the force modality in sim A using SAC-Lagrangian, and then we fine-tune it with vision and force in sim B with SAC-Lagrangian. 

For training the multimodal policy and the force residual policy, to encourage the policy to exert low force to the human, we add an additional penalty term to the reward when the force is above the threshold. For the multimodal safe RL baseline, the force is treated as the cost for SAC-Lagrangian.  

All methods are evaluated in sim B, to demonstrate the ability of each method to transfer to new dynamics. 
We provide additional details of these baselines on our project website.
We train all multimodal baselines till convergence, which usually require a magnitude more data than FCVP. 

\noindent\textbf{Evaluation Metrics:}
The first evaluation metric is the 
\textit{Arm Dressed Ratio}, which is the ratio between the dressed arm distance and the real arm length. A ratio of 1 means that the arm is fully dressed; 0 means that the arm is not dressed at all.  This metric is used to measure the dressing performance and has been used in related work~\cite{Wang2023One}. 
The second metric is the \textit{Average Force Violation}, which is computed as $\frac1T{\sum_{t=1}^T\max(0, f_t - \tau)}$, in which $f_t$ is the measured force at time step $t$ and $\tau$ is the maximal force threshold.

\noindent\textbf{Results:}
Figure~\ref{fig:sim2sim_results} presents the performances of all methods. As shown on the left subplot, FCVP achieves the best trade-off between the dressed ratio and the force violation amount: it achieves the lowest force violation, and the third highest arm dressed ratio. All other baselines either have large force violations (Force Residual Policy, Vision Only, Force Only), or perform poorly on the dressing task, as indicated by a low Arm Dressed Ratio (Multimodal policy, Multimodal Safe RL, Force only). 
As shown in the right table of Figure~\ref{fig:sim2sim_results}, most other multimodal learning baselines require much more training data in sim B, which serves as a proxy of the real world. This is because they require policy fine-tuning via RL in sim B. In contrast, FCVP learns a dynamics model in sim B, which is a supervised learning problem, thus being much more sample-efficient to learn. 
Overall, these results show that both vision and force information are needed to achieve high dressing performance while being safe; further, our approach of incorporating vision and force in FCVP not only achieves better performance, but also requires much less training data in the ``real world" (sim B) than other multimodal learning methods. 
\rebuttal{Box plots of the force distributions for all methods are provided on our project website for further visualization.}

\noindent\textbf{Ablation study.}
\rebuttal{We first investigate how the number of past force measurements, denoted as $N$, affects the performance of FCVP. We test $N=3, 5, 7$ and find that a larger $N$ leads to a decrease in force violations but also a reduced dressed ratio. We use $N=5$, which achieves a good trade off between these two objectives.
We also study the impact of including past actions as part of the input for the force dynamics model. We test including $0, 1, 3, 5$ steps of past actions, and find the performance varies minimally across these different lengths. The best performance is achieved when the model does not include past actions as part of its input. Please refer to our project website for further details on these ablation studies. 
}

\subsection{Real-World Human Study}
\label{sec:human_study}

\begin{figure}[t]
    \centering
\includegraphics[width=.5\textwidth]{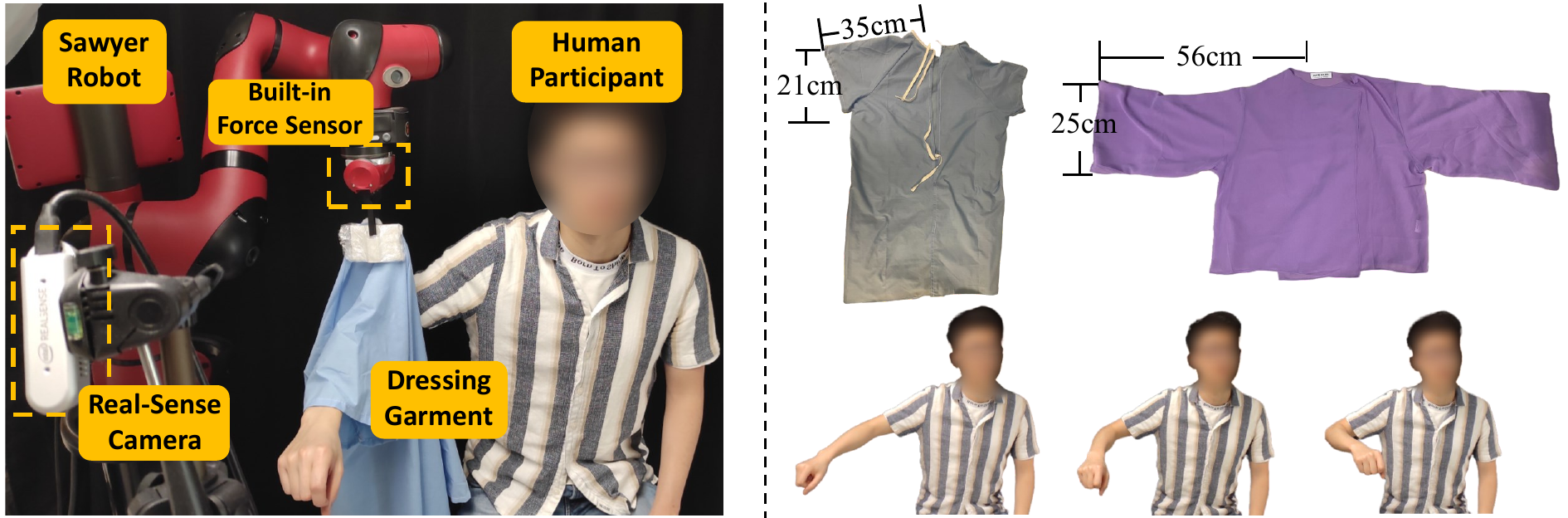}
\vspace{-0.25in}
    \caption{Left: Human study setup.  Right: Poses and Garments that we test in the human study.}
\label{fig:real_world_pose_and_garment}
\vspace{-0.2in}
\end{figure}

\noindent\textbf{Experimental Setup:} 
\label{sec:human_study}
Figure~\ref{fig:real_world_pose_and_garment} shows the setup for our real-world human study. We use the Sawyer robot for executing the dressing task, and measured the force at the Sawyer robot’s end-effector (wrist) using Sawyer's built-in force sensing. The robot movement (action) is the delta translation and rotation of the end-effector, and is executed using the Sawyer’s built-in IK solver and an impedance controller.
We use a single Intel RealSense D435i camera to capture the point cloud observations of the scene. 
We compare our method with the following two baselines: 

\textbf{Vision Only}~\cite{Wang2023One} chooses the action with the highest probability under the vision-based policy. 

\textbf{Vision with Random Actions} samples the action from the vision-based policy with probability $p$, and uniformly randomly from $[-1, 1]^{|A|}$ with probability $1-p$, where $|A|$ is the dimension of the action space (6 in our case). 

We compare to the ``Vision with Random Actions" baseline as it is also used for collecting data for training the force dynamics model, as mentioned in section~\ref{sec:learning_force}. 
We set $p=0.1$ and the force threshold $\tau$ to be 5.4 Newtons in our experiments, which is an empirical value based on the force distributions of successful and safe dressing trials, and we think this value would be comfortable for the participants during the dressing process.
We do not compare to the \textit{Force Only} baseline because it performs poorly in terms of the dressed ratios even in simulation (see table in Figure~\ref{fig:sim2sim_results}). 
We also do not compare to other multimodal learning baselines, as they all require extensive amounts of training data to converge (as shown in the sim2sim transfer experiments in Figure~\ref{fig:sim2sim_results}), which is prohibitive to collect in the real world. 

In addition to the evaluation metrics described in section~\ref{sec:sim2sim_experiments}, we present participants with 7-point Likert items that range from 1=`Strongly Disagree' to 7=`Strongly Agree' with the following statements: 1. ``The robot successfully dressed the garment onto my arm''; 2. ``The force the robot applied to me during dressing was appropriate''; and 3. ``The dressing process was comfortable for me.''

\noindent\textbf{Human Study Procedure:}
\label{sec:human_study_procedure}
We recruit 10 participants, comprising 4 males, 4 females, 1 non-binary individual, and 1 participant who chooses not to disclose their gender identity. The age of the participants spans from 19 to 36.  
We test each participants with two garments: a short-sleeve hospital gown and a long-sleeve purple cardigan. 
These two garments differ in geometry (see Fig~\ref{fig:real_world_pose_and_garment} for sleeve lengths and widths), elasticity and roughness (the purple cardigan is more elastic), and mass (the hospital gown is heavier). 
We test each participant with two different arm poses, randomly chosen from three poses. The poses and garments we use are shown in Figure~\ref{fig:real_world_pose_and_garment}. 
We evaluate each of the three methods for two trials on each pose-garment combination, resulting in a total of 24 trials per participant. We also randomize the ordering of the test poses, garments, and methods. Participants are asked to hold their arm steady throughout each trial. We run each trial for a fixed number of time steps, unless the participant asks to stop or the measured force is above a safety threshold (15 Newtons). We train a single force dynamics model and test it on all these 10 participants. 

\noindent\textbf{Force dynamics model training:} Before evaluating FCVP in the human study, we need to capture real-world force data and train a force dynamics model. 
We capture force data from a separate 11-participant study, using a similar procedure as noted above. The 11 participants consist of 6 males and 5 females with ages ranging from 22 to 50. 
For each participant, we first run the Vision with Random Actions baseline for 8 trials, 2 trials for each combination of 2 arm poses and 2 garments. Using the force data collected in these 8 trials, we train a force dynamics model on this participant. We then run our method (FCVP) with the trained force dynamics model, as well as the Vision Only baseline (order randomized) for 8 trials on the same arm poses and garments. 
This results in 24 dressing trajectories for each of the 11 participants. 
By using 3 different methods, we are able to enlarge the distribution of the captured force data, which is beneficial for training a single generalized force dynamics model.
These 3 methods used for data collection are optimized for either completing the dressing task (Vision Only baseline and Vision with Random Action baseline), or reducing the force (FCVP), thus they are all safer compared to random trajectories, reducing the safety risk posed to the participants during data collection. 
We ask participants to hold their arm steady throughout each trial when collecting the force data to ensure the accuracy of the collected data.
We use the force data captured during the 264 dressing trials on these 11 participants to train a single generalized force dynamics model, and evaluate it in another human study with 10 new participants (gender and age distributions as described above). 
Note that the 10 new participants we test in the evaluation of FCVP with the generalized force dynamics model are all different from the 11 participants used for collecting the force data. More details of the study procedure for both human studies can be found in our project website. 

\begin{table}[t]
    \centering
    \scriptsize
        \caption{\footnotesize Quantitative results of the human study. FCVP not only achieves higher arm dressed ratio, but also has significantly lower force violations.
    }
    \vspace{-0.06in}
    \begin{tabular}{ccc}
    \toprule
   & {\begin{tabular}[c]{@{}c@{}} Arm  \\ Dressed Ratio $\uparrow$ \end{tabular}}  & \multicolumn{1}{c}{\begin{tabular}[c]{@{}c@{}}Average \\ Force Violation (N) $\downarrow$ \end{tabular}}  \\ \midrule
     FCVP (Ours)   & $\mathbf{0.81 \pm 0.21}$ & $\mathbf{0.089}$  \\ 
     Vision Only~\cite{Wang2023One}    &  $0.71 \pm 0.17$ & $0.39$  \\ 
     Vision w/ Random Action  & $0.71 \pm 0.18$  & $0.34$ 
     \\ 
     \bottomrule
    \end{tabular}
    \label{tab:real_result}
    \vspace{-0.2in}
\end{table}

\begin{figure*}[t]
    \centering
    \includegraphics[width=.87\textwidth]{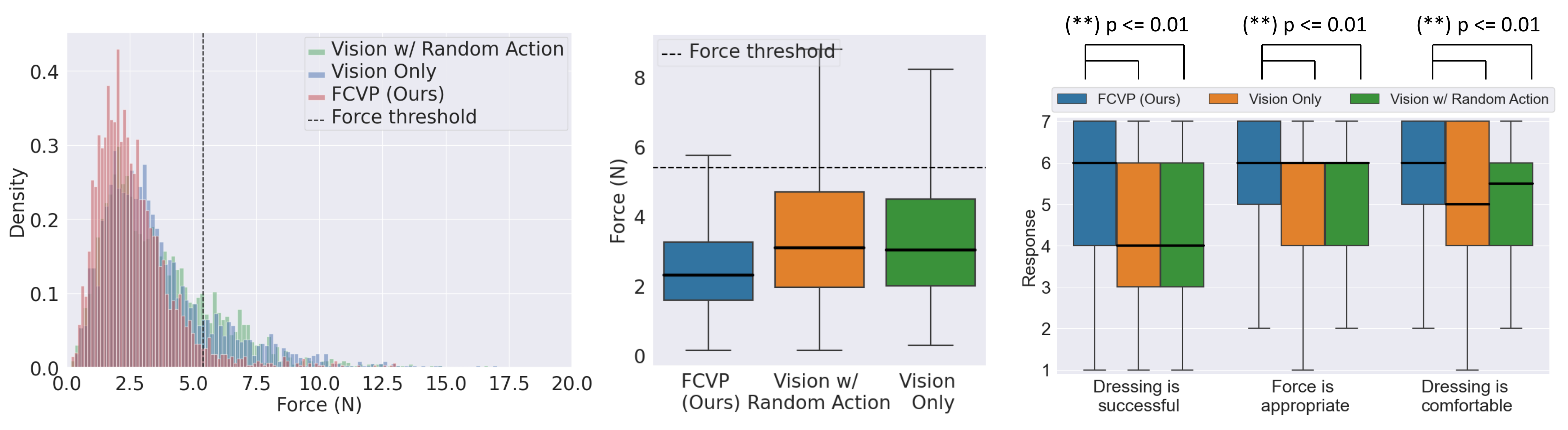}
    \vspace{-0.15in}
    \caption{Left and middle: Density plot and box plot of the force distributions on all participants in the human study. The dashed black line represents the force threshold. Our method greatly reduces the force violation compared to the baselines. Right: Likert item responses from all 10 participants. FCVP achieves statistically significant differences from both baselines with higher reported scores for all 3 Likert items.
    }
    \label{fig:real_result}
    \vspace{-0.15in}
\end{figure*}

\noindent\textbf{Human Study Results:}
Videos of the dressing trials are available on our project website. 
Table~\ref{tab:real_result} compares the results of all methods, averaged over all 10 participants. As shown, our \algo~has significantly lower force violation compared to both baselines. Furthermore, it achieves higher arm dressed ratios. Due to sim2real transfer gaps, the vision-based policy often deviates to a state where the garment gets caught on the person. This state usually does not occur in simulation, as the simulated garments are often more elastic with lower frictional coefficients than those in the real world, due to limited simulation fidelity. 
By constraining the amount of force applied to the person, the force dynamics model guides the vision-based policy to avoid such scenarios, and as a result, the final dressing performance is higher (see Figure~\ref{fig:intro_example} for an example).
This same factor might affect other multimodal learning baseline methods as well, potentially leading to a similar discrepancy between their sim2sim and sim2real performance. Still, the performance of the other baseline methods would likely be lower than that of our method (FCVP) in the real world, since their performance in simulation is quite poor (see Fig~\ref{fig:sim2sim_results}). Further, due to the large amount of training data (thousands of dressing trajectories) needed for fine-tuning other multimodal learning methods (see the last column of the table in Fig~\ref{fig:sim2sim_results}), it is often impractical to train and test these methods in the sim2real setting. In contrast, our method is able to efficiently learn a dynamics model from just 264 real world trajectories. Thus, our method is both higher performing and more practical than the other approaches.
We note that the performance of the Vision Only baseline is lower than reported by~\cite{Wang2023One}. This is due to us testing on only a subset of the garments and poses (the most difficult ones) among those tested in~\cite{Wang2023One}.  

Figure~\ref{fig:real_result} shows the density (left) and box (middle) plots of the force distributions of all participants. As shown, most of the forces exerted by FCVP are below the force threshold, while the other two baselines violate the threshold more frequently, demonstrating the strong effectiveness of our method for reducing the force violation. 
Figure~\ref{fig:real_result} (right) also reports the Likert Item responses of all participants. On average, FCVP achieves a median score of 6 for all three Likert items, meaning that the participants ``Agree'' that the robot successfully dressed the garment onto their arm, ``Agree'' the force the robot applied to them during dressing is appropriate, and ``Agree'' that the dressing process was comfortable for them. 
The medians of FCVP are higher than or the same as both baselines for all 3 Likert items.  
We conduct a Wilcoxon signed rank test to test if 
the distribution of the paired difference in scores between two methods is different from a distribution symmetric about zero. For all three Likert items and both baselines,
we obtain a p-value smaller than 0.01 ($p < 0.01$), i.e., we find a statistically significant difference between FCVP and the two baselines, and the median of the difference is greater than zero.
We note that our human study is performed on a younger age distribution than the group that will likely need assistive dressing. However, as our experiments have shown, the dynamics model is able to generalize to new participants within the same age distribution; thus we believe that if it is trained with an older age distribution, it should perform well on that older age distribution as well. We hope to verify this in future studies. 

\noindent\textbf{Generalization of the force dynamics model.}
\rebuttal{
As the input to the force dynamics model is the partial point cloud of the scene, which captures the shape and size of the human arm, the force dynamics model should be able to generalize to the shape and size of the arm within the training distribution. 
In our human study, the size and shape of the arms of the 10 evaluation participants are different to those of the 11 training participants. 
The average prediction error (L1 norm) of the force dynamics model is 0.00631N on the 11 training participants, and 0.0478N on the 10 evaluation participants. Despite this train-eval gap, the evaluation prediction error is still small ($< 0.1$N), and the force dynamics model still proves to be useful in reducing force violations, as shown in Table~\ref{tab:real_result}. These results indicate that the force dynamics model can generalize reasonably well to the shape and size of the human arms. We believe the gap can be narrowed in future work by regularization, or collecting more data. 
}

\rebuttal{
We also analyze the generalization of the force dynamics model to different physical properties of the garments, such as roughness and elasticity, in simulation. We create 50 garments with the same geometry but different elasticity, by randomly sampling the spring coefficients of the garments within a range of $[0.3, 1.5]$. We then train a force dynamics model on 40 garments and test on the remaining 10. The force dynamics model generalizes well to the garments with unseen elasticity: the average force violation increased slightly from 2.21 to 4.13 simulation units from training garments to unseen garments (a vision-based policy trained on the 10 unseen garments has a force violation of 38.8 simulation units). 
More details of these experiments can be found on our project website. 
}

\noindent\textbf{System analysis.} 
\rebuttal{
There can be cases where there is no action whose predicted force is below the safety threshold.
We present the ratio of the number of timesteps where this situation occurs to the total number of timesteps. Findings from our human study indicate that this ratio is low, at $4.028\%$, showing that such cases are uncommon and generally don't impact dressing performance. 
The ratio could be further lowered by sampling more actions when solving the constrained optimization problem. 
The average inference time taken to solve the optimization problem in the real-world experiments is 0.065 seconds per timestep. Each dressing trial usually lasts between 40 and 60 seconds.
}

\vspace{-0.05in}
\subsection{Limitations}
One limitation is that our method assumes the person holds a static pose and that the robot has already grasped the garment. Prior works have introduced new sensors and control strategies specifically targeting at relaxing these assumptions~\cite{gao2015user, zhang2019probabilistic,  erickson2022characterizing}, which could be applied in conjunction with our work. 
Additionally, in our experiments, we note that FCVP still has some forces that exceed the threshold. There are two reasons for this: First, there may be minor errors in the learned force dynamics model's prediction.
An action can be predicted to be below the threshold, but actually applies more force when executed.
This issue can be mitigated by collecting more data for training the force dynamics model to make it more accurate, or by adding a buffer to the force threshold to account for prediction inaccuracies. 
Second, because our method uses random shooting to solve the constrained optimization problem, the solutions may not satisfy the constraint if all sampled actions are infeasible. This would result in some forces above the threshold. 
This issue can be alleviated by sampling more actions until a feasible action is found, or by using alternative optimization algorithms to solve the constrained optimization problem. 
\rebuttal{At last, we request all participants to wear short-sleeve T-shirts during dressing. The properties of the cloth they wear, such as friction and elasticity, could affect the forces applied to the users during the dressing process and affect the ground-truth force labels used to train the force dynamics model. We have not tested if the force dynamics model generalizes to other clothing the users wear, such as pulling a jacket over a long-sleeve shirt, which we leave as interesting future work.
}

%% file: sections/conclusion.tex
\section{Conclusion}
In this paper, we propose a new method to leverage both vision and force modalities for robot-assisted dressing, based on which we build a system that combines these modalities to ensure task progress and low applied forces for safety. We learn a vision-based policy via reinforcement learning in simulation across diverse people, poses, and garments. We train a force dynamics model directly in the real world to achieve safety and overcome inaccuracies in simulated force sensing with deformable garments. Our system combines the vision-based policy and the force model via a constrained optimization problem to find actions that progress the dressing process without applying excessive force to the person. We evaluate our system in simulation and in a real-world human study with 10 participants and 240 trials, demonstrating that it greatly outperforms prior baselines. 

%% file: main.bbl
\begin{thebibliography}{10}
\providecommand{\url}[1]{#1}
\csname url@rmstyle\endcsname
\providecommand{\newblock}{\relax}
\providecommand{\bibinfo}[2]{#2}
\providecommand\BIBentrySTDinterwordspacing{\spaceskip=0pt\relax}
\providecommand\BIBentryALTinterwordstretchfactor{4}
\providecommand\BIBentryALTinterwordspacing{\spaceskip=\fontdimen2\font plus
\BIBentryALTinterwordstretchfactor\fontdimen3\font minus
  \fontdimen4\font\relax}
\providecommand\BIBforeignlanguage[2]{{%
\expandafter\ifx\csname l@#1\endcsname\relax
\typeout{** WARNING: IEEEtran.bst: No hyphenation pattern has been}%
\typeout{** loaded for the language `#1'. Using the pattern for}%
\typeout{** the default language instead.}%
\else
\language=\csname l@#1\endcsname
\fi
#2}}

\bibitem{harris2019long}
L.~D. Harris-Kojetin, M.~Sengupta, J.~P. Lendon, V.~Rome, R.~Valverde, and
  C.~Caffrey, ``Long-term care providers and services users in the united
  states, 2015-2016.'' 2019.

\bibitem{Wang2023One}
Y.~Wang, Z.~Sun, Z.~Erickson, and D.~Held, ``One policy to dress them all:
  Learning to dress people with diverse poses and garments,'' in
  \emph{Robotics: Science and Systems (RSS)}, 2023.

\bibitem{zhang2022learning}
F.~Zhang and Y.~Demiris, ``Learning garment manipulation policies toward
  robot-assisted dressing,'' \emph{Science robotics}, vol.~7, no.~65, p.
  eabm6010, 2022.

\bibitem{kapusta2019personalized}
A.~Kapusta, Z.~Erickson, H.~M. Clever, W.~Yu, C.~K. Liu, G.~Turk, and C.~C.
  Kemp, ``Personalized collaborative plans for robot-assisted dressing via
  optimization and simulation,'' \emph{Autonomous Robots}, vol.~43, no.~8, pp.
  2183--2207, 2019.

\bibitem{gao2016iterative}
Y.~Gao, H.~J. Chang, and Y.~Demiris, ``Iterative path optimisation for
  personalised dressing assistance using vision and force information,'' in
  \emph{2016 IEEE/RSJ international conference on intelligent robots and
  systems (IROS)}.\hskip 1em plus 0.5em minus 0.4em\relax IEEE, 2016, pp.
  4398--4403.

\bibitem{gao2015user}
Y.~Gao, J.~Chang, and Y.~Demiris, ``User modelling for personalised dressing
  assistance by humanoid robots,'' in \emph{2015 IEEE/RSJ International
  Conference on Intelligent Robots and Systems (IROS)}.\hskip 1em plus 0.5em
  minus 0.4em\relax IEEE, 2015, pp. 1840--1845.

\bibitem{erickson2018deep}
Z.~Erickson, H.~M. Clever, G.~Turk, C.~K. Liu, and C.~C. Kemp, ``Deep haptic
  model predictive control for robot-assisted dressing,'' in \emph{2018 IEEE
  international conference on robotics and automation (ICRA)}.\hskip 1em plus
  0.5em minus 0.4em\relax IEEE, 2018, pp. 4437--4444.

\bibitem{zhang2019probabilistic}
F.~Zhang, A.~Cully, and Y.~Demiris, ``Probabilistic real-time user posture
  tracking for personalized robot-assisted dressing,'' \emph{IEEE Transactions
  on Robotics}, vol.~35, no.~4, pp. 873--888, 2019.

\bibitem{zhang2017personalized}
F.~Zhang, A.~Cully, and Y.~Dimiris, ``Personalized robot-assisted dressing
  using user modeling in latent spaces,'' in \emph{2017 IEEE/RSJ International
  Conference on Intelligent Robots and Systems (IROS)}.\hskip 1em plus 0.5em
  minus 0.4em\relax IEEE, 2017, pp. 3603--3610.

\bibitem{zhang2020learning}
F.~Zhang and Y.~Demiris, ``Learning grasping points for garment manipulation in
  robot-assisted dressing,'' in \emph{2020 IEEE International Conference on
  Robotics and Automation (ICRA)}.\hskip 1em plus 0.5em minus 0.4em\relax IEEE,
  2020, pp. 9114--9120.

\bibitem{tamei2011reinforcement}
T.~Tamei, T.~Matsubara, A.~Rai, and T.~Shibata, ``Reinforcement learning of
  clothing assistance with a dual-arm robot,'' in \emph{2011 11th IEEE-RAS
  International Conference on Humanoid Robots}.\hskip 1em plus 0.5em minus
  0.4em\relax IEEE, 2011, pp. 733--738.

\bibitem{koganti2014real}
N.~Koganti, T.~Tamei, T.~Matsubara, and T.~Shibata, ``Real-time estimation of
  human-cloth topological relationship using depth sensor for robotic clothing
  assistance,'' in \emph{The 23rd IEEE international symposium on robot and
  human interactive communication}.\hskip 1em plus 0.5em minus 0.4em\relax
  IEEE, 2014, pp. 124--129.

\bibitem{pignat2017learning}
E.~Pignat and S.~Calinon, ``Learning adaptive dressing assistance from human
  demonstration,'' \emph{Robotics and Autonomous Systems}, vol.~93, pp. 61--75,
  2017.

\bibitem{erickson2017does}
Z.~Erickson, A.~Clegg, W.~Yu, G.~Turk, C.~K. Liu, and C.~C. Kemp, ``What does
  the person feel? learning to infer applied forces during robot-assisted
  dressing,'' in \emph{2017 IEEE International Conference on Robotics and
  Automation (ICRA)}.\hskip 1em plus 0.5em minus 0.4em\relax IEEE, 2017, pp.
  6058--6065.

\bibitem{yu2017haptic}
W.~Yu, A.~Kapusta, J.~Tan, C.~C. Kemp, G.~Turk, and C.~K. Liu, ``Haptic
  simulation for robot-assisted dressing,'' in \emph{2017 IEEE international
  conference on robotics and automation (ICRA)}.\hskip 1em plus 0.5em minus
  0.4em\relax IEEE, 2017, pp. 6044--6051.

\bibitem{kapusta2016data}
A.~Kapusta, W.~Yu, T.~Bhattacharjee, C.~K. Liu, G.~Turk, and C.~C. Kemp,
  ``Data-driven haptic perception for robot-assisted dressing,'' in \emph{2016
  25th IEEE international symposium on robot and human interactive
  communication (RO-MAN)}.\hskip 1em plus 0.5em minus 0.4em\relax IEEE, 2016,
  pp. 451--458.

\bibitem{wang2022visual}
Y.~Wang, D.~Held, and Z.~Erickson, ``Visual haptic reasoning: Estimating
  contact forces by observing deformable object interactions,'' \emph{IEEE
  Robotics and Automation Letters}, vol.~7, no.~4, pp. 11\,426--11\,433, 2022.

\bibitem{sunil2023visuotactile}
N.~Sunil, S.~Wang, Y.~She, E.~Adelson, and A.~R. Garcia, ``Visuotactile
  affordances for cloth manipulation with local control,'' in \emph{Conference
  on Robot Learning}.\hskip 1em plus 0.5em minus 0.4em\relax PMLR, 2023, pp.
  1596--1606.

\bibitem{lee2019making}
M.~A. Lee, Y.~Zhu, K.~Srinivasan, P.~Shah, S.~Savarese, L.~Fei-Fei, A.~Garg,
  and J.~Bohg, ``Making sense of vision and touch: Self-supervised learning of
  multimodal representations for contact-rich tasks,'' in \emph{2019
  International Conference on Robotics and Automation (ICRA)}.\hskip 1em plus
  0.5em minus 0.4em\relax IEEE, 2019, pp. 8943--8950.

\bibitem{watkins2019multi}
D.~Watkins-Valls, J.~Varley, and P.~Allen, ``Multi-modal geometric learning for
  grasping and manipulation,'' in \emph{2019 International conference on
  robotics and automation (ICRA)}.\hskip 1em plus 0.5em minus 0.4em\relax IEEE,
  2019, pp. 7339--7345.

\bibitem{calandra2018more}
R.~Calandra, A.~Owens, D.~Jayaraman, J.~Lin, W.~Yuan, J.~Malik, E.~H. Adelson,
  and S.~Levine, ``More than a feeling: Learning to grasp and regrasp using
  vision and touch,'' \emph{IEEE Robotics and Automation Letters}, vol.~3,
  no.~4, pp. 3300--3307, 2018.

\bibitem{wi2022virdo}
Y.~Wi, P.~Florence, A.~Zeng, and N.~Fazeli, ``Virdo: Visio-tactile implicit
  representations of deformable objects,'' in \emph{2022 International
  Conference on Robotics and Automation (ICRA)}.\hskip 1em plus 0.5em minus
  0.4em\relax IEEE, 2022, pp. 3583--3590.

\bibitem{sundaresan2022learning}
P.~Sundaresan, S.~Belkhale, and D.~Sadigh, ``Learning visuo-haptic skewering
  strategies for robot-assisted feeding,'' in \emph{6th Annual Conference on
  Robot Learning}, 2022.

\bibitem{du2022play}
M.~Du, O.~Y. Lee, S.~Nair, and C.~Finn, ``Play it by ear: Learning skills
  amidst occlusion through audio-visual imitation learning,'' \emph{arXiv
  preprint arXiv:2205.14850}, 2022.

\bibitem{yang2017deep}
X.~Yang, P.~Ramesh, R.~Chitta, S.~Madhvanath, E.~A. Bernal, and J.~Luo, ``Deep
  multimodal representation learning from temporal data,'' in \emph{Proceedings
  of the IEEE conference on computer vision and pattern recognition}, 2017, pp.
  5447--5455.

\bibitem{li2022see}
H.~Li, Y.~Zhang, J.~Zhu, S.~Wang, M.~A. Lee, H.~Xu, E.~Adelson, L.~Fei-Fei,
  R.~Gao, and J.~Wu, ``See, hear, and feel: Smart sensory fusion for robotic
  manipulation,'' \emph{arXiv preprint arXiv:2212.03858}, 2022.

\bibitem{beltran2020variable}
C.~C. Beltran-Hernandez, D.~Petit, I.~G. Ramirez-Alpizar, and K.~Harada,
  ``Variable compliance control for robotic peg-in-hole assembly: A
  deep-reinforcement-learning approach,'' \emph{Applied Sciences}, vol.~10,
  no.~19, p. 6923, 2020.

\bibitem{garcia2015comprehensive}
J.~Garc{\i}a and F.~Fern{\'a}ndez, ``A comprehensive survey on safe
  reinforcement learning,'' \emph{Journal of Machine Learning Research},
  vol.~16, no.~1, pp. 1437--1480, 2015.

\bibitem{liu2022constrained}
Z.~Liu, Z.~Cen, V.~Isenbaev, W.~Liu, S.~Wu, B.~Li, and D.~Zhao, ``Constrained
  variational policy optimization for safe reinforcement learning,'' in
  \emph{International Conference on Machine Learning}.\hskip 1em plus 0.5em
  minus 0.4em\relax PMLR, 2022, pp. 13\,644--13\,668.

\bibitem{erickson2022characterizing}
Z.~Erickson, H.~M. Clever, V.~Gangaram, E.~Xing, G.~Turk, C.~K. Liu, and C.~C.
  Kemp, ``Characterizing multidimensional capacitive servoing for physical
  human-robot interaction,'' \emph{IEEE Transactions on Robotics (T-RO)}, 2022.

\bibitem{haarnoja2018soft}
T.~Haarnoja, A.~Zhou, P.~Abbeel, and S.~Levine, ``Soft actor-critic: Off-policy
  maximum entropy deep reinforcement learning with a stochastic actor,'' in
  \emph{International conference on machine learning}.\hskip 1em plus 0.5em
  minus 0.4em\relax PMLR, 2018, pp. 1861--1870.

\bibitem{qi2017pointnet++}
C.~R. Qi, L.~Yi, H.~Su, and L.~J. Guibas, ``Pointnet++: Deep hierarchical
  feature learning on point sets in a metric space,'' \emph{Advances in neural
  information processing systems}, vol.~30, 2017.

\bibitem{qi2017pointnet}
C.~R. Qi, H.~Su, K.~Mo, and L.~J. Guibas, ``Pointnet: Deep learning on point
  sets for 3d classification and segmentation,'' in \emph{Proceedings of the
  IEEE conference on computer vision and pattern recognition}, 2017, pp.
  652--660.

\bibitem{macklin2019non}
M.~Macklin, K.~Erleben, M.~M{\"u}ller, N.~Chentanez, S.~Jeschke, and
  V.~Makoviychuk, ``Non-smooth newton methods for deformable multi-body
  dynamics,'' \emph{ACM Transactions on Graphics (TOG)}, vol.~38, no.~5, pp.
  1--20, 2019.

\bibitem{lin2021softgym}
X.~Lin, Y.~Wang, J.~Olkin, and D.~Held, ``Softgym: Benchmarking deep
  reinforcement learning for deformable object manipulation,'' in
  \emph{Conference on Robot Learning}.\hskip 1em plus 0.5em minus 0.4em\relax
  PMLR, 2021, pp. 432--448.

\bibitem{SMPL-X:2019}
G.~Pavlakos, V.~Choutas, N.~Ghorbani, T.~Bolkart, A.~A.~A. Osman, D.~Tzionas,
  and M.~J. Black, ``Expressive body capture: {3D} hands, face, and body from a
  single image,'' in \emph{Proceedings IEEE Conf. on Computer Vision and
  Pattern Recognition (CVPR)}, 2019, pp. 10\,975--10\,985.

\bibitem{bertiche2020cloth3d}
H.~Bertiche, M.~Madadi, and S.~Escalera, ``Cloth3d: clothed 3d humans,'' in
  \emph{European Conference on Computer Vision}.\hskip 1em plus 0.5em minus
  0.4em\relax Springer, 2020, pp. 344--359.

\end{thebibliography}
